\pdfoutput=1

\documentclass[11pt]{article}

\usepackage[final]{acl}

\usepackage{times}
\usepackage{latexsym}
\usepackage[T1]{fontenc}

\usepackage[utf8]{inputenc}

\usepackage{microtype}

\usepackage{inconsolata}

\usepackage{braket}
\usepackage{graphicx}
\usepackage{amsmath}
\usepackage{amssymb}
\usepackage{booktabs}
\usepackage{multirow}
\usepackage{multicol}
\usepackage{tabularx}
\usepackage{arydshln}
\usepackage{makecell}
\usepackage[T1]{fontenc}
\usepackage{pifont}
\usepackage{xspace}

\newcommand\blfootnote[1]{%
  \begingroup
  \renewcommand\thefootnote{}\footnote{#1}%
  \addtocounter{footnote}{-1}%
  \endgroup
}

\title{Let's Go Real Talk: \\ Spoken Dialogue Model for Face-to-Face Conversation}

\author{
\begin{tabular}{c}
Se Jin Park$^*$ \quad Chae Won Kim$^*$ \quad Hyeongseop Rha \quad Minsu Kim\\ \quad Joanna Hong \quad Jeong Hun Yeo \quad Yong Man Ro$^\dagger$ 
\end{tabular} \\ 
Integrated Vision and Language Lab, KAIST \\ 
\tt\small  
\begin{tabular}{c}
\texttt{\{jinny960812, chaewonkim, ryool\_1832, sedne246, ymro\}@kaist.ac.kr}\\ \quad \texttt{ms.k@ieee.org} \quad \texttt{joanna2587@gmail.com}  
\end{tabular}
}

\begin{document}
\maketitle
\blfootnote{$^*$Equal contribution. $^\dagger$Corresponding author. This work was supported by the National Research Foundation of Korea (NRF) grant funded by the Korea government (MSIT) (No.~NRF-2022R1A2C2005529) and Institute of Information \& communications Technology Planning \& Evaluation (IITP) grant funded by the Korea government (MSIT) (No.2022-0-00124, Development of Artificial Intelligence Technology for Self-Improving Competency-Aware Learning Capabilities).}
\begin{abstract}
In this paper, we introduce a novel Face-to-Face spoken dialogue model. It processes audio-visual speech from user input and generates audio-visual speech as the response, marking the initial step towards creating an avatar chatbot system without relying on intermediate text. To this end, we newly introduce MultiDialog, the first large-scale multimodal (\ie, audio and visual) spoken dialogue corpus containing 340 hours of approximately 9,000 dialogues, recorded based on the open domain dialogue dataset, TopicalChat. The MultiDialog contains parallel audio-visual recordings of conversation partners acting according to the given script with emotion annotations, which we expect to open up research opportunities in multimodal synthesis. Our Face-to-Face spoken dialogue model incorporates a textually pretrained large language model and adapts it into the audio-visual spoken dialogue domain by incorporating speech-text joint pretraining. Through extensive experiments, we validate the effectiveness of our model in facilitating a face-to-face conversation. Demo and data are available at \url{https://multidialog.github.io} and \url{https://huggingface.co/datasets/IVLLab/MultiDialog}, respectively. 
\end{abstract}

\begin{table*}
	\renewcommand{\arraystretch}{1.3}
	\renewcommand{\tabcolsep}{2.5mm}
\centering
\resizebox{0.85\linewidth}{!}{
\begin{tabular}{cccccccc}
\Xhline{3\arrayrulewidth}
\textbf{Dataset} & \textbf{\# Dialogues}& \textbf{\# Turns} &  \textbf{Length (hrs)} & \textbf{Audio} & \textbf{Text} & \textbf{Video} & \textbf{Emotion}  \\
\hline
IEMOCAP \cite{busso2008iemocap} & 151 & 10,039 & 12 & \ding{51} & \ding{51} & \ding{51} & \ding{51}  \\
DSTC2 \cite{henderson2014second} & 1,612 & 23,354 & 32 & \ding{51} & \ding{51} & \ding{55} & \ding{55} \\
MELD \cite{poria2018meld} &  1,433 & 13,000 & 13.7 & \ding{51} & \ding{55} & \ding{51} & \ding{51}  \\
DailyTalk \cite{lee2023dailytalk} & 2,514 & 23,774 & 21.7 & \ding{51} & \ding{51} & \ding{55} & \ding{55}  \\
Expresso \cite{nguyen2023expresso} & 391 & 2,400 & 47 & \ding{51} & \ding{51} & \ding{55} & \ding{51}  \\
SpokenWOZ \cite{si2023spokenwoz} & 5,700 & 203,074 & 249 & \ding{51} & \ding{51} & \ding{55} & \ding{55}  \\
\textbf{MultiDialog} & 8,733 & 187,859 & 340 & \ding{51} & \ding{51} & \ding{51} & \ding{51}   \\
\bottomrule
\Xhline{3\arrayrulewidth}
\end{tabular}}
\vspace{-0.1cm}
\caption{Comparison of MultiDialog dataset with publicly available multimodal dialogue datasets.}
\label{table:1}
\vspace{-0.3cm}
\end{table*}

\section{Introduction}
Spoken Dialogue System (SDS), often referred to as a conversational agent, engages in natural speech conversations with humans by recognizing speech from user input and providing contextually appropriate and accurate responses with speech. With spoken language as the primary interface, it has numerous applications for human-computer interactions such as customer service and voice assistants. 

However, when people communicate face-to-face, we utilize not only audio but also visual information of the conversing partner to process spoken words and non-verbal cues (\ie, facial expressions, gestures, and emotions) \cite{petridis2018end,hong2023watch}. This multimodal information enhances understanding of the speech content and the speaker's intent. Furthermore, having a visual counterpart to audio can simulate a real face-to-face conversation experience, making the user feel more connected and engaged.


In this paper, we explore an audio-visual spoken dialogue system to facilitate direct face-to-face conversation for the first time. Central to the development of dialogue systems is the large amount of high-quality dialogue data. Current dialogue systems are predominantly text-based, driven by the abundance of text dialogue datasets \cite{lowe2015ubuntu, DailyDialog, zhang2018personalizing, EmpatheticDialogues, budzianowski2018multiwoz,  zhou2018dataset, reddy2019coqa, lamberthuggingface, ding2023enhancing, kopf2023openassistant}. Recently, several audio dialogue datasets have been released \cite{lee2023dailytalk, si2023spokenwoz, nguyen2023expresso} which augment existing text dialogue data \cite{DailyDialog, budzianowski2018multiwoz} with speech. However, those with visual components remain limited in scale, comprising less than 15 hours in total \cite{busso2008iemocap,poria2018meld}. Addressing this data gap, we introduce MultiDialog, the first large-scale audio-visual spoken dialogue corpus. It consists of 340 hours of audio-visual recordings of approximately 9,000 dialogues, derived from open-domain text dialogue dataset, TopicalChat \cite{gopalakrishnan2023topical} which is an extensive multi-turn dialogue corpus collected from real conversations covering 9 broad topics. The proposed MultiDialog consists of emotion annotations for each utterance and simultaneous recordings of both the listener and the speaker, presenting opportunities for diverse research; from face-to-face dialogue system to talking face synthesis \cite{park2022synctalkface, zhang2023sadtalker}, listener's face synthesis \cite{song2023emotional,zhou2023interactive}, and emotion-conditioned face synthesis \cite{goyal2023emotionally}.

Based on the MultiDialog dataset, we propose the first audio-visual spoken dialogue model that can directly process audio-visual speech as user input and generate audio-visual speech as the output response. Motivated by the recent success of the direct spoken dialogue model using discretized speech tokens \cite{nguyen2023generative, zhang2023speechgpt}, we introduce audio-visual (AV) speech tokens extracted by quantizing audio-visual speech features from a self-supervised model \cite{shi2022learning}. Utilizing the AV speech tokens as pseudo texts, we integrate AV speech into a pretrained large-language model (LLM) \cite{zhang2022opt} through joint speech-text pretraining. 
The response is also returned in AV speech tokens, which are synthesized into a talking face video as the output for direct interaction with the system. 

Our contributions are in three folds: 
(1) We introduce the first direct Face-to-Face dialogue model which processes multimodal speech from user input and generates multimodal speech as the output response, facilitating a face-to-face conversation system. 
(2) To build a face-to-face dialogue system, we propose the first large-scale multimodal (\ie, audio, visual, and text) dialogue corpus, MultiDialog consisting of 340 hours of approximately 9,000 audio-visual conversation streams. 
(3) We demonstrate that joint speech-text pretraining leveraging a pre-trained large language model improves upon direct initialization in retaining knowledge of the original large language model. 


\begin{table*}
	\renewcommand{\arraystretch}{1.2}
	\renewcommand{\tabcolsep}{4mm}
\centering
\resizebox{0.85\linewidth}{!}{
\begin{tabular}{p{4cm}ccccc:c}
\Xhline{3\arrayrulewidth}
\textbf{MultiDialog} & \textbf{Train} & \textbf{Valid Freq} & \textbf{Valid Rare} & \textbf{Test Freq} & \textbf{Test Rare} & \textbf{Total}   \\
\hline
\# dialogues & 7,011 & 448 & 443 & 450 & 381 & 8,733 \\
\# utterance  & 151,645 & 8,516 & 9,556 & 9,811 & 8,331 & 187,859  \\
avg \# utterance/dialogue  & 21.63 & 19.01 & 21.57 & 21.80 & 21.87 & 21.51 \\
avg length/utterance (s)  & 6.50 & 6.23 & 6.40 & 6.99 & 6.49 & 6.51 \\
avg length/dialogue (min)  & 2.34 & 1.97 & 2.28 & 2.54 & 2.36 & 2.33 \\
total length (hr)  & 273.93 & 14.74 & 17.00 & 19.04 & 15.01 & 339.71 \\
\bottomrule
\Xhline{3\arrayrulewidth}
\end{tabular}}
\vspace{-0.1cm}
\caption{Datailed statistics of MultiDialog}
\label{table:2}
\vspace{-0.2cm}
\end{table*}

\section{Related Work}
\subsection{Spoken Dialogue Dataset}
In recent years, the development of speech dialogue datasets has played a pivotal role in understanding human behavior and building spoken dialogue systems that emulate real-life conversations. Early speech datasets focus on analyzing human behavior such as emotion and intent in speech, establishing the foundation for spoken dialogue systems. IEMOCAP \cite{busso2008iemocap} and MELD \cite{poria2018meld}, comprising audio and video recordings of dialogues, are designed to study emotional dynamics in conversations. In addition to understanding emotions, DSTC2 \cite{henderson2014second} presents telephone-based speech dialogues for dialogue state tracking to predict user’s goals. Building upon datasets that study human behavior in speech, recent spoken dialogue datasets were built to model realistic dialogue systems. Expresso \cite{nguyen2023expresso} introduces speech dialogues spanning 26 expressive styles for natural speech synthesis. DailyTalk \cite{lee2023dailytalk} and SpokenWOZ \cite{si2023spokenwoz} datasets introduce speech-text conversations for spoken dialogues. While existing works have contributed to advancing spoken conversation systems, dialogue datasets are limited in scale and solely consist of audio and text, thereby constraining the development of audio-visual spoken dialogue systems incorporating visual cues. To address these limitations, we expand the spoken dialogue in scale and to the visual modality, and introduce MultiDialog, a large-scale multimodal spoken dialogue dataset. A summary of existing multimodal dialogue datasets and MultiDialog is shown in  Table~\ref{table:1}.

\subsection{Spoken Dialogue Models} 
Audio Language Model, driven by transformer-based architecture, has made remarkable strides in speech processing. By treating continuous speech as a discrete set of representations, speech can be effectively modeled as text, allowing the application of Natural Language Processing (NLP) techniques. While it has made notable progress in speech synthesis \cite{lakhotia2021generative, borsos2023audiolm, wang2023neural, hassid2023textually, nachmani2023lms}, speech translation \cite{barrault2023seamlessm4t, dong2023polyvoice, rubenstein2023audiopalm}, and speech recognition \cite{wang2023viola}, spoken dialogue system is a relatively unexplored field of research due to the scarcity of spoken dialogue datasets. Several works made an effort to tackle data issues by leveraging the power of large language models (LLMs). SpeechGPT \cite{zhang2023speechgpt} first converts speech into discrete speech tokens, and then designs a three-stage training pipeline on paired speech data, speech instruction data, and chain-of-modality instruction data. AudioGPT \cite{huang2023audiogpt} instructs LLMs to generate commands for controlling external tools before inputting them into the LLMs. d-GSLM \cite{nguyen2023generative} models two-channel conversations to produce natural turn-taking conversations. 

There are Multimodal Large Language Models (MM-LLM) \cite{wu2023next, gong2023multimodal} capable of processing both visual input and output. However, they are visual grounding dialogue systems that use visual information as supplementary for tasks such as image captioning and image editing. In contrast, we aim to build an audio-visual spoken dialogue system (\ie, facial movement related to the speech) to enhance the understanding of speech content and enrich the communication experience, emulating a real face-to-face conversation.

\section{MultiDialog Dataset}
\subsection{Preparation}
To obtain audio-visual recordings of dialogues, we gathered 12 fluent English speakers, with varying gender, age, and nationality. The participants, aged 20 to 30, came from six different countries, with six female and six male actors, as shown in Appendix~\ref{sec:participant}. We derived dialogue scripts from the open-domain dialogue dataset, TopicalChat \cite{gopalakrishnan2023topical} which is a rich knowledge-grounded dataset collected from real human-human conversations. It spans eight broad topics including fashion, politics, books, sports, general entertainment, music, science \& technology, and movies. It is annotated for eight emotions: Disgusted, Angry, Fearful, Happy, Sad, Surprised, Neutral, and Curious to dive deeper. The conversation partners don't have explicitly defined roles as `speaker' or `listener' so they interact naturally similar to how people engage in real-world conversations. Due to the topical variety, emotion annotation, and representation of natural human conversations, we chose TopicalChat as the foundation for constructing the multimodal dialogue dataset. 

\subsection{Recording}
Data was recorded in a professional recording studio with a green screen and minimal background noise, shown in Appendix~\ref{sec:dataset1}. During a recording session, two conversation partners sat side-by-side and were recorded with a separate camera and a microphone. The camera position was adjusted according to the individual's height to capture the upper body, starting from the shoulders. The participants were asked to act according to a given script conveying the desired emotion annotation for each utterance. We specifically provided detailed instructions for visual and audio cues based on the Facial Action Coding System \cite{ekman1978facial} and tone \cite{gangamohan2016analysis} for each emotion as follows:
\begin{itemize}
    \setlength\itemsep{-0.5em} 
    \item \textbf{Neutral}: normal resting face, emotionless, speak still with natural information.
    \item \textbf{Happy}: lip corner puller, cheek raiser, lips parts, speak cheerfully in a higher tone.
    \item \textbf{Sad}: drooping upper eyelids, slight pulling down of lips corners, speak in a sad, lower tone.
    \item \textbf{Fearful}: eyebrows raised and pulled together, eye pulled open, speak in a soft and low tone.
    \item \textbf{Surprise}: eyebrows raised, eyes wide open, mouth open wider, speak excitedly with high tone.
    \item \textbf{Disgusted}: eyebrows lowered and pulled together, nose wrinkled, cheek raised, upper lip raised, speak in a normal tone with disgusted intonation.
    \item \textbf{Anger}: eyebrows lowered and pulled together, eyes glare, speak powerfully with high tone.
\end{itemize}
For recordings, we combined the emotion labels `Neutral' and `Curious to dive deeper' into a single label `Neutral' due to the lack of visually apparent difference between the two. In addition to the instructions, we displayed sample images on the screen so that the actors could mimic the facial expressions corresponding to the emotion. 
Moreover, when the turn passes to another participant, they naturally react while listening. Participants were instructed to press a button to proceed to the next utterance, which recorded the start and end times of each turn for post-processing. The audio streams were recorded in a mono WAV format at 48kHz and the video streams in full HD at 30fps. 

\subsection{Post-Processing} 
To refine the data, we had an annotator go through the audio-visual recordings to check if there were any misalignments between the audio and visual streams. We asked the annotator to manually adjust the misalignments by sliding the start time. Additionally, we filtered out recordings that were missing either audio or visual streams. Then, we segmented the recordings into conversations and turns based on the recorded timesteps of each turn. As a result, the post-processed MultiDialog dataset consists of approximately 340 hours of audio-visual videos of 9,000 dialogues between 6 pairs of conversation partners. The final statistics of our dataset are shown in Table \ref{table:2}. Furthermore, we release a gold emotion dialogue subset selected based on rigorous annotation evaluation. Please refer to the Appendix~\ref{sec:goldemotion} for more details.

\begin{figure*}[t]
	\begin{minipage}[b]{\linewidth}
		\centering		\centerline{\includegraphics[width=11cm]{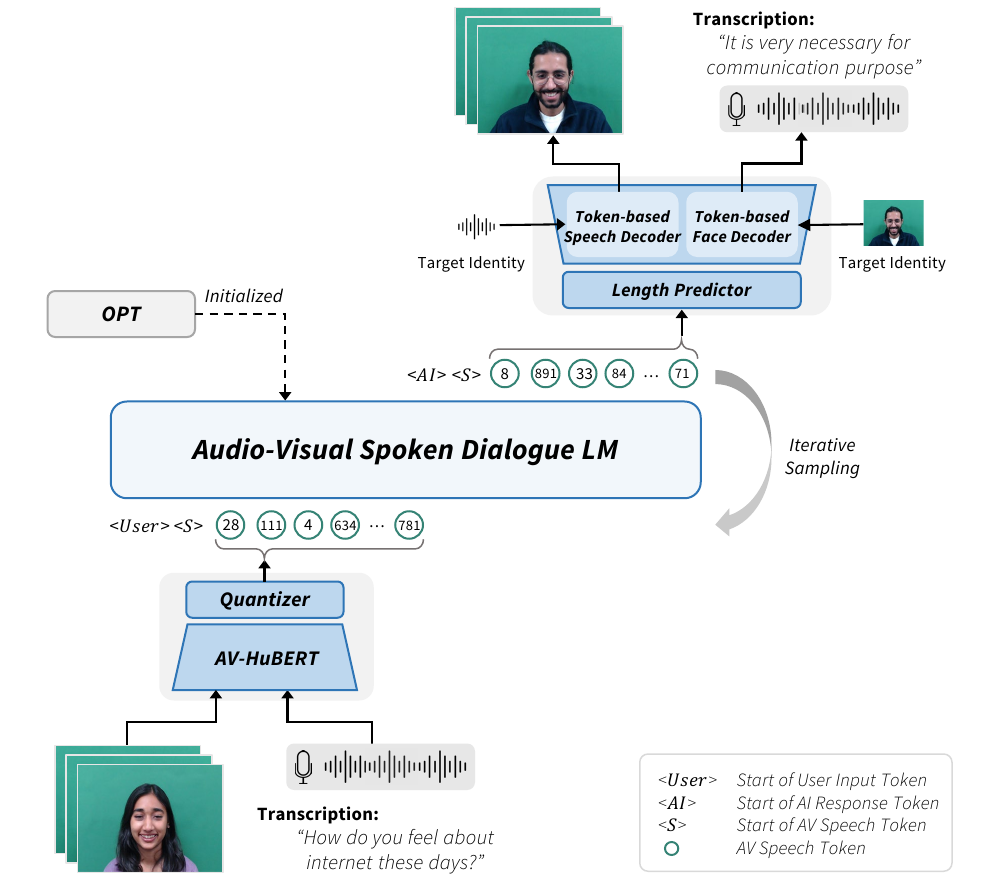}}
	\end{minipage}
	\caption{Overview of the proposed framework for multimodal spoken dialogue language modeling. With the AV speech tokens as the pseudo-texts, it can process audio-visual face video from the user input and generate corresponding response as audio-visual face video.}
	\label{fig:1}
  \vspace{-0.4cm}
\end{figure*}

\section{Audio-Visual Spoken Dialogue System}
Based on the proposed MultiDialog dataset, we introduce an audio-visual spoken dialogue system that directly understands the audio-visual of the user's face video and generates appropriate responses with audio-visual face video. It consists of three main parts: 1) Encoding audio-visual speech into discrete representations, namely audio-visual (AV) speech tokens. 2) Conducting multimodal spoken dialogue language modeling using the AV speech tokens as pseudo texts. 3) Projecting the output AV speech tokens into the audio and visual space for direct face-to-face dialogue.   

\subsection{Audio-Visual Speech Encoding}
By integrating both audio and visual modalities, we can improve the dialogue system's understanding of the speech content. This is because speech not only comprises auditory signals but also visual cues from the movements of the speaker's mouth. This visual information complements auditory signals, particularly in noisy environments, resulting in more robust performance \cite{afouras2018deep}. 

To this end, we adopt a unified approach to model both the audio and visual of talking face input into audio-visual speech tokens. Inspired by the recent success of utilizing discrete speech tokens extracted from self-supervised speech models \cite{schneider2019wav2vec, baevski2020wav2vec, hsu2021hubert, chung2021w2v, babu2021xls} in speech processing \cite{lakhotia2021generative, lee2021textless, maiti2023voxtlm, kim2023many}, we tokenize the audio and visual streams into audio-visual speech tokens (\textit{a.k.a.} AV speech tokens). 
Specifically, we employ one of the multimodal speech models, AV-HuBERT \cite{shi2022learning}, a state-of-the-art self-supervised framework for understanding speech by both seeing and hearing. It is trained on raw audio-visual face videos to predict discrete clusters from speech \cite{hassid2023textually}. The audio-visual speech features are extracted and quantized into discrete tokens as in \cite{lakhotia2021generative, popuri2022enhanced, kim2024multilingual}. By combining the visual cues and the auditory information, the audio-visual speech tokens extract both linguistic and phonetic information. Then, we treat the AV speech tokens as pseudo text to train our Audio-Visual Spoken Dialogue LM.

\subsection{Audio-Visual Spoken Dialogue Language Modeling} 
As shown in Fig.~\ref{fig:1}, our audio-visual spoken dialogue language model is trained with the AV speech tokens on our MultiDialog dataset. 
Previous work \cite{hassid2023textually} showed that initializing a speech language model with a textually pretrained language model (LLM) leads to better performance and faster convergence. Accordingly, we use a pretrained LLM, OPT-1.3B \cite{zhang2022opt} to initialize our model and combine the vocabulary of AV speech tokens with the original text vocabulary, as in \cite{zhang2023speechgpt, nachmani2023lms, maiti2023voxtlm}. This allows us to jointly model the probability of both AV speech tokens and text tokens $t$, where the loss can be represented as, 
\begin{align}
\mathcal{L}=-\sum_{i=1}^{N}\log p(t_i \mid t_1, ... , t_{i-1}), 
\end{align}
which is the negative log-likelihood of predicting the next token in the sequence of length $N$ tokens.

\begin{figure}[t!]
	\begin{minipage}[b]{\linewidth}
		\centering		\centerline{\includegraphics[width=8cm]{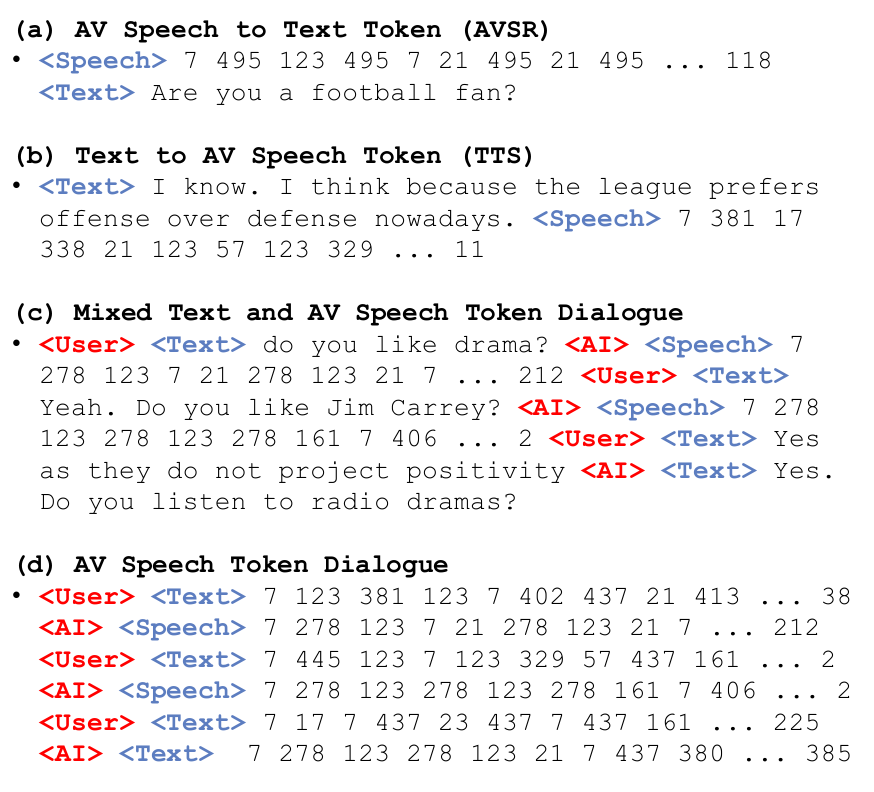}}
	\end{minipage}
	\caption{Constructed data based on the MultiDialog dataset used for training the audio-visual speech dialogue model. (a-c) are joint pretraining of the audio-visual speech and text tokens and (d) is used to finetune the model.}
	\label{fig:2}
  \vspace{-0.4cm}
\end{figure}

Motivated by the joint speech-text training used in speech processing tasks such as speech translation, audio speech recognition, and text-to-speech synthesis \cite{cheng2023mu, maiti2023voxtlm, dong2023polyvoice, wang2023viola}, we newly introduce a joint speech-text pre-training scheme tailored for spoken dialogue language modeling. In our setting, each dialogue $D = [T_1^{ai}, T_1^{user}, T_2^{ai}, T_2^{user}, \ldots, T_k^{ai}, T_k^{user}]$ consists of $k$ rounds of turns $T$ between two speakers which we randomly designate as the AI and the User. The goal of this pre-training is to effectively transform the text-based LLM into the AV speech token-based LLM, enabling it to produce relevant AV speech responses from the AI side given a conversation context. It proceeds in the following two stages: 

\textbf{The first stage} is instructing the LLM to interpret and generate AV speech tokens. We segment the dialogue into turns $T$ and prepare paired AV speech tokens $T_{\text{AV}}$ and text tokens $T_{\text{Text}}$. We then concatenate the pair with their respective modality prefix tokens, $\textrm{<speech>}$ and $\textrm{<text>}$, to indicate the beginning of AV speech and text tokens. Adding the reversed order of concatenation, we construct both audio-visual speech recognition (AVSR) and text-to-speech generation (TTS) training objectives as shown in Fig.~\ref{fig:2}(a) and (b), where the loss functions can be respectively represented as: 
\begin{align}
\mathcal{L}_{\text{AVSR}} &= \sum_{i=1}^{N} -\log p(\ T_{\text{AV}}^i \mid T_{\text{AV}}^{<i}, T_{\text{Text}})\\
\mathcal{L}_{\text{TTS}} &= \sum_{i=1}^{N} -\log p( T_{\text{Text}}^i \mid T_{\text{Text}}^{<i}, T_{\text{AV}}).
\end{align}
We omitted the prefix tokens for conciseness. Only the embedding layer and the projection layer are trained in the first stage, which guides the LLM to understand and generate AV speech tokens while fully retaining the given LLM knowledge needed for dialogue generation. 

\textbf{The second stage} is jointly learning the text and AV speech token-based dialogue. We select either one of the speakers as the AI which the model aims to predict and indicate the start of the response with additional speaker prefix tokens, $\textrm{<User>}$ and $\textrm{<AI>}$. The speaker prefix token is followed by a modality prefix token, $\textrm{<Speech>}$ and $\textrm{<Text>}$, to indicate whether the utterance is in AV speech or text. 
The loss function for dialogue language modeling is: 
\begin{align}
\mathcal{L}_{\text{dialog}} = \sum_{k=1}^{K} \sum_{n=1}^{N_k} -\log p(T_{k}^{ai,n} \mid T_{k}^{ai,<n}, T_{<k}), 
\end{align}
where $K$ is the total number of rounds, $N_k$ is the number of tokens in the k-th round,  $T_{k}^{ai,n}$ is the n-th token from the AI in the k-th round, $T_{k}^{ai,<n}$ denotes all previous tokens from the AI within the same round k, and $T_{<k}$ is all prior tokens in previous rounds. Note that we dropped the prefix tokens in the equation for brevity. During the pretraining, we utilize a balanced mix of the AV speech tokens and text which allows the model to utilize both token knowledge to generate dialogue response as in Fig.~\ref{fig:2}(c). Then, we later finetune on pure AV speech token-based dialogue as in Fig.~\ref{fig:2}(d) for real-time face-to-face interaction. This progressive shift helps the model to gradually adapt to AV speech tokens without compromising the quality of dialogue generation of the text-based LLM.

\begin{figure}[t!]
	\begin{minipage}[b]{\linewidth}
		\centering		\centerline{\includegraphics[width=8cm]{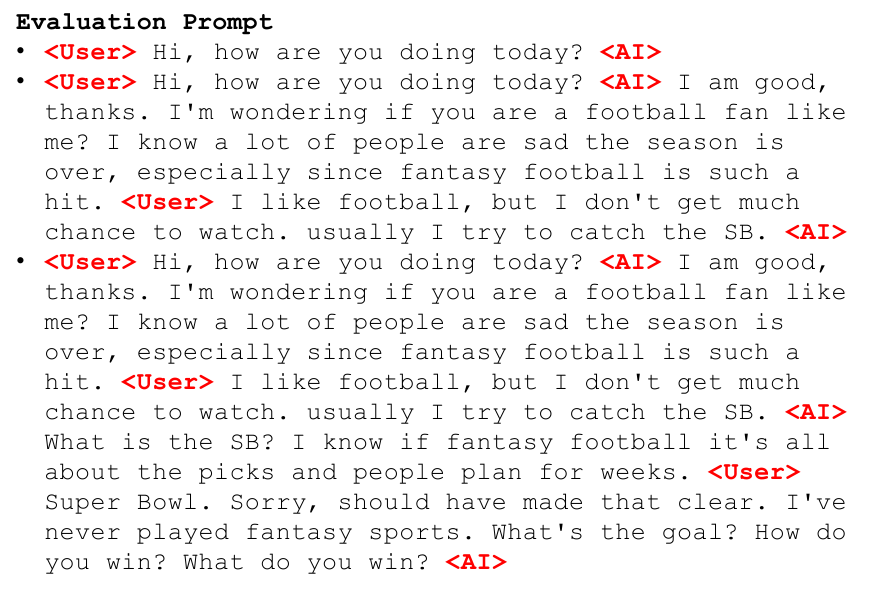}}
	\end{minipage}
	\caption{Evaluation prompt of multimodal dialogue language modeling. It is written in text for illustration but the actual prompt is given as audio and visual. }
	\label{fig:3}
  \vspace{-0.4cm}
\end{figure}

\subsection{Audio-Visual Generation}
The generated AV speech tokens are projected to audio and visual to generate the response as a talking face video. As shown in Fig.~\ref{fig:1}, the audio-visual generator consists of a length predictor, a token-based speech decoder, and a token-based face decoder. Since our language model is trained with duplicate reduced AV speech tokens, we train a length predictor to first restore them back to their original length. The token-based speech decoder and token-based face decoder are adapted from an off-the-shelf audio generator \cite{kong2020hifi} and a talking face generator \cite{prajwal2020lip} respectively, where we train them to process AV speech tokens as the input instead of raw audio. Additionally, we incorporate speaker identity information by extracting the speaker embedding \cite{jia2018transfer} from a target identity sample audio. Also, the target identity's face and pose prior are utilized as in \cite{prajwal2020lip}, to enable the generation of talking face video with desired identity. 

\section{Experimental Setup}
\subsection{Evaluation Metrics}
We evaluate the semantic quality and the generation quality of both audio and video. For the semantic quality, we first generate transcriptions from the synthesized audio-visual output using an off-the-shelf ASR model \cite{shi2022learning}, and employ standard metrics used for text-based dialogue generation: log-perplexity (PPL), BLEU, METEOR, F1, D-1, and D-2. The log-perplexity is calculated using Dialo-GPT model \cite{zhang2019dialogpt} and it is calculated for each utterance and averaged across the test set. To measure the generation quality of video, we adopt metrics used for TFG. This includes Fréchet Inception Distance (FID) \cite{heusel2017gans} to measure visual quality, and LSE-C and LSE-D \cite{prajwal2020lip} to measure the audio-visual synchronization. To evaluate the acoustic quality, we compute speaker similarity (SIM) between the given target sample and generated speech using the WavLM-Base model for speaker verification \cite{chen2022wavlm}. Please refer to the appendices for a detailed explanation of each metric. 

\begin{table*}
	\renewcommand{\arraystretch}{1.3}
	\renewcommand{\tabcolsep}{2.8mm}
\centering
\resizebox{0.99\linewidth}{!}{
\begin{tabular}{lccccccccc}
\Xhline{3\arrayrulewidth}
\multirow{2}{*}{\makecell{\textbf{Method}}} 
& \multirow{2}{*}{\makecell{\textbf{Input}\\ \textbf{Modality}}} 
& \multirow{2}{*}{\makecell{\textbf{Output}\\ \textbf{Modality}}} 
&
\multicolumn{7}{c}{\textbf{Semantic Evaluation}} \\ \cmidrule(lr){4-10}  

& &  & \textbf{PPL} $\downarrow$ & \textbf{BLEU} $\uparrow$ & \textbf{METEOR} $\uparrow$ & \textbf{F1} $\uparrow$ & \textbf{D-1 } $\uparrow$ & \textbf{D-2} $\uparrow$ 
\\ \hline
\multicolumn{4}{l}{\quad $\bullet$ \textbf{\textit{Ground Truth}}} \\ 
GT AV Speech Token &  -- & -- & 1054.643 & 76.326 & 0.565 & 0.474 & 0.947 & 0.996  \\

\multicolumn{4}{l}{\quad $\bullet$ \textbf{\textit{Cascaded System}}} \\ 
 AVSR + LM + TTS + TFG &  AV & AV & 1157.586 & 47.287 & 0.075 & 0.100 & 0.959 & 0.977   \\

\multicolumn{4}{l}{\quad $\bullet$ \textbf{\textit{Spoken Dialogue System}}} \\ 
SpeechGPT \cite{zhang2023speechgpt} & A & A & 930.401 & 20.536 & 0.064 & 0.054 & 0.743 & 0.876 \\
d-GSLM \cite{nguyen2023generative} & A & A  & 1085.265 & 8.197 & 0.065 & 0.064 & 0.883 & 0.876 \\

\hdashline 
\multicolumn{4}{l}{\quad $\bullet$ \textbf{\textit{Audio-Visual Spoken Dialogue System}}} \\ 
Scratch & AV & AV & 1898.864 & 13.305 & 0.058 & 0.064 & 0.945 & 0.955 \\
+ LLM initialized & AV & AV & 1237.757 & 17.098 & 0.059 & 0.058 & 0.936 & 0.963 \\
+ AVSR/TTS Pretraining & AV & AV & 1068.904 & 22.090 & 0.062 & 0.066 & 0.943 & 0.965 \\
+ Mixed Text-AV Speech Pretraining & AV & AV & 1248.001 & 24.094 & 0.063 & 0.065 & 0.945 & 0.957  \\

\Xhline{3\arrayrulewidth}
\end{tabular}}
\vspace{-0.1cm}
\caption{Comparison of the semantic quality between state-of-the-art spoken dialogue systems in MultiDialog. Note that our proposed method is the only method that supports both audio and visual at the input and output of the dialogue system without relying on intermediate text. }
\label{table:4}
\vspace{-0.2cm}
\end{table*}

\subsection{Implementation Details}
To encode AV speech tokens, we crop the video into the mouth region of size 96$\times$96 using a face detector \cite{deng2020retinaface} and a facial landmark detector \cite{bulat2017far}, and resample the audio to 16kHz. We take English-trained AV-HuBERT \cite{shi2022learning} and finetune it to predict corresponding target clusters from HuBERT tokenizer \cite{hassid2023textually} which operates at 25Hz with 500 clusters. We train it for 100k steps on 6 A6000 GPUs with a maximum token length of 2,000. 

We initialize the model with a pre-trained language model, OPT-1.3B \cite{zhang2022opt}. We first pretrain the input embedding layer and the projection layer on AVSR and TTS objectives for 200K steps. Then, we continue training the entire model on a mixture of text and AV speech token dialogue for 5K steps, followed by finetuning for additional 3K steps on AV speech token dialogue only. We use a max token length of 700 on 4 A6000 GPUs. 

The audio-visual generator is trained using ground truth AV speech tokens. The token-based speech decoder and length predictor are jointly trained for 450K steps with a batch size of 32. For training the AV token-based face decoder, we employ the reprogramming strategy in \cite{choi2023reprogramming} and train an adapter layer consisting of two layers of transformer encoder to bridge between the AV speech tokens and the corresponding audio features of the TFG model \cite{prajwal2020lip}. This allows to leverage the face generation capabilities of the pretrained TFG model without further finetuning the generator and can be applied to any other TFG models. It is trained for 250K steps with a batch size of 256. We additionally incorporate a face enhancer \cite{wang2021gfpgan} to upsample the generated face video into high resolution.

\subsection{Baselines}
Since there is no previous method that can directly perform audio-visual spoken dialogue synthesis, we compare with the recently proposed spoken dialogue systems, Speech-GPT \cite{zhang2023speechgpt} and d-GSLM \cite{nguyen2023generative}. They support only audio speech at both input and output. Additionally, we build a cascade system by integrating a series of off-the-shelf pre-trained models: AVSR \cite{anwar2023muavic}, LM \cite{tang2022mvp}, TTS \cite{casanova2022yourtts}, and TFG \cite{prajwal2020lip}. Please note the objective of the comparisons with the cascaded method is not to achieve state-of-the-art performance, but rather to assess the extent to which the performance of the proposed system can be attained through the direct strategy. For a fair comparison, we finetune SpeechGPT and d-GSLM on our MultiDialog dataset and we use a dialogue language model \cite{tang2022mvp} trained on TopicalChat as the LM of the cascade system.

\section{Results}
\subsection{Semantic Evaluation}
To accurately assess the semantic quality of the generated response, we employ the evaluation strategy used for text-based dialogue language models. We conduct evaluations on the test set of MultiDialog, where the model is prompted to sequentially generate a response for each turn in the conversations. Sample evaluation prompts are illustrated in Figure \ref{fig:3}. The generated response is then transcribed into text and compared against the ground truth response to evaluate its semantic quality. As shown in Table \ref{table:4}, compared with the state-of-the-art spoken dialogue systems, SpeechGPT \cite{zhang2023speechgpt} and d-GSLM \cite{nguyen2023generative}, our proposed method performs the best in BLEU, D-1, and D-2 which demonstrates that our method can generate contextually coherent and diverse response. SpeechGPT has the highest PPL because it is trained on an extensive amount of speech data and PEFT-finetuned \cite{hu2021lora} on the MultiDialog, which allows it to generate more fluent speech but fails to match with the reference response as indicated by the lower BLEU score. Also, it requires generating text transcription of the input to generate the response in text first. Notably, our proposed method stands as the first approach to directly recognize and generate response in both audio and visual speech video, without requiring intermediate text generation.

\subsection{Ablation on the Pretraining Scheme}
We analyze the pretraining scheme used for our audio-visual spoken dialogue model in the lower section of Table \ref{table:4}. The results demonstrate that initializing the model with a textually pretrained LLM yields improved semantic quality, which is further enhanced by AVSR/TTS pretraining. Simply training the embedding layer and projection layer to predict corresponding AV speech tokens and text tokens improves the response. When further incorporating mixed text-AV speech token pretraining, we observe an overall enhancement in semantic quality, validating the effectiveness of gradually adapting the AV speech tokens to the LLM. Yet, there is a slight decrease in the PPL score, which we attribute to the model's increased complexity and adaptability to multimodal inputs. 

\subsection{Audio and Visual Evaluation}
\begin{table}
\renewcommand{\arraystretch}{1.3}
\renewcommand{\tabcolsep}{0.9mm}
\centering
\resizebox{0.99\linewidth}{!}{
\begin{tabular}{lcccc}
\Xhline{3\arrayrulewidth}
\textbf{Method} & FID $\uparrow$ & LSE-C $\uparrow$ & LSE-D $\downarrow$ & SIM $\uparrow$  \\ \hline
\multicolumn{3}{l}{\quad $\bullet$ \textbf{\textit{Cascade System}}} \\ 
{AVSR + LM + TTS + TFG} & 30.581 & 7.041 & 7.640 & 0.433 \\
\hdashline
\multicolumn{3}{l}{\quad $\bullet$ \textbf{\textit{Spoken Dialogue System}}} \\ 
{SpeechGPT \cite{zhang2023speechgpt}} & - & - & - & 0.194 \\
{d-GSLM \cite{nguyen2023generative}} & - & - & - & 0.211   \\
\hdashline
\multicolumn{3}{l}{\quad $\bullet$ \textbf{\textit{Audio-Visual Spoken Dialogue System}}} \\ 
\textbf{Proposed} & 30.323 & 7.298 & 7.390 & 0.624 \\
\Xhline{3\arrayrulewidth}
\end{tabular}}
\vspace{-0.1cm}
\caption{Evaluation of the audio and visual generation quality. Note that we evaluate the reconstructed audio and visual output of randomly selected 300 videos from the test set of MultiDialog. \label{table:5} }
\vspace{-0.3cm}
\end{table}

\begin{figure*}[t]
	\begin{minipage}[b]{\linewidth}
		\centering		\centerline{\includegraphics[width=17cm]{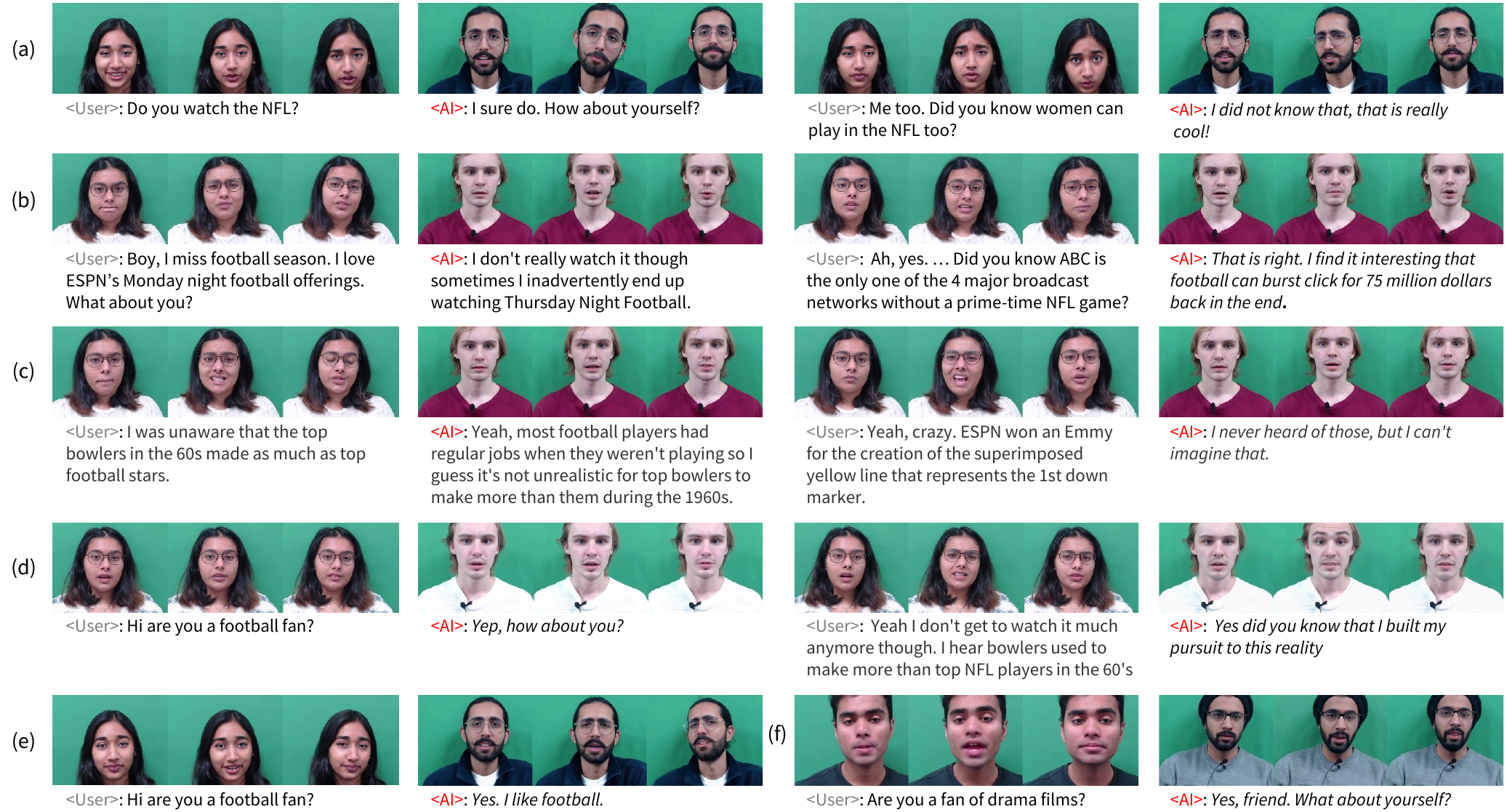}}
	\end{minipage}
        \vspace{-0.5cm}
	\caption{Audio-visual dialogue generation results of the proposed method, where the last turn is the generated audio-visual response. Note that we have randomly sampled three video frames from each turn for illustration. (a-d) are conversations with four turns and (e-f) are with two turns, The generated responses are in italics and we provide ASR transcriptions below.}
	\label{fig:4}
  \vspace{-0.3cm}
\end{figure*}
We evaluate the audio and visual generation quality in Table \ref{table:5}. In terms of speaker voice similarity (SIM), our proposed method not only outperforms the cascaded system but also surpasses spoken dialogue systems. This demonstrates the effectiveness of our AV token-based speech decoder, enriched with speaker embedding, to retain the speaker information from the reference video. When assessing visual quality, we compared it with the cascaded system that uses the same TFG model \cite{prajwal2020lip} as ours. While our FID score is comparable, our approach exhibits superior audio-visual synchronization, due to the utilization of discretized audio-visual tokens, which provide clearer alignment between the audio and visual components than raw audio.

In Figure \ref{fig:4}, we show the generated audio-visual response between the two partners along with transcriptions generated with ASR \cite{shi2022learning}. Given a conversation context, our model generates the next response that is contextually coherent and adequate. For example, in Figure \ref{fig:4} (a), it answers the question asked by the user in the previous turn and responds accordingly about the chatting topic, NFL. Also, it successfully synthesizes the speech-relevant movements of the reference face to generate a seamless talking face video. Please refer to the demo for more demonstrations. 

\subsection{Robustness to Acoustic Noise}
\begin{table}
	\renewcommand{\arraystretch}{1.3}
	\renewcommand{\tabcolsep}{2.4mm}
\centering
\resizebox{0.999\linewidth}{!}{
\begin{tabular}{lcccccc}
\Xhline{3\arrayrulewidth}
\multirow{2}{*}{\textbf{Method}} & \multirow{2}{*}{\makecell{\textbf{Input}\\\textbf{Modality}}}
& \multicolumn{5}{c}{\centering \textbf{SNR (dB)}} \\ 
\cmidrule(lr){3-6} 
& & -5 & 0 & 5 & clean
\\
\hline 
\multirow{2}{*}{\makecell{\textbf{Proposed}}} & $A$ & 11.340 & 14.751 & 21.143 & 23.089 \\
& $AV$ & 13.853 & 18.144 & 21.186 & 24.094 \\
\Xhline{3\arrayrulewidth}
\end{tabular}}
\vspace{-0.1cm}
\caption{\label{table:6} Dialogue response generation performance (BLEU) with different input modalities under acoustic noise corruption with different SNR levels (dB).}
\vspace{-0.2cm}
\end{table}
In Table \ref{table:6}, we analyze the effectiveness of incorporating additional visual modality into the dialogue system. Following \cite{shi2022learning}, we corrupt the input speech with random noise of varying SNR levels (-5, 0, 5, and clean). Compared with audio-only input, audio-visual input enhances the robustness of the system as indicated by less degradation of the performance under noise. This is because the visual modality which is not affected by acoustic noise can complement the missing information in the audio modality to better recognize the speech content and output response. It further demonstrates that our system is applicable for real-world use in unstable speech input scenarios. 

\section{Conclusion and Limitation}
We introduce a novel face-to-face spoken dialogue model that directly processes audio-visual speech from the user input and generates audio-visual speech response. This is the first step toward creating a talking face avatar chatbot system, without intermediate text in the generation process. In addition, we release MultiDialog, the largest multimodal dialogue dataset to date with tri-modality (\ie, audio, visual, and text) spoken dialogue data. As it is an extensive dataset that captures real human-human conversation covering broad topics, we believe it brings diverse research opportunities for multimodal synthesis, ranging from talking face synthesis to multimodal dialogue language modeling. 

One limitation of our work is that, although the dataset includes emotion labels for each utterance, we have not utilized these labels yet. We plan to address this in future research by integrating emotion recognition from users' facial expressions to generate more emotion-aware responses, both in speech content and nuances of generation. Also, since our data provides parallel recordings of the speaker and the listener, we can simultaneously model the generation of both faces for more spontaneous and natural conversation. 

\bibliography{main}

\newpage
\appendix
\section{MultiDialog Dataset}
\label{sec:dataset}

\subsection{Dataset Statistics}
Table~\ref{table:2} shows detailed statistics of MultiDialog. MultiDialog consists of 9,920 human-human conversations, 106,624 turns, 218,248 utterances, totalling to approximately 340 hours of audiovisual dialogue data. A single dialogue contains multiple turns, where each turn includes two utterances. An utterance is an instance of speech by one person followed by silence or another person speaking. In our dataset, a conversation averaged 11.0 turns, 21.9 utterances, 140.2 seconds in length. 12 speakers were paired to record an average of 826.7 dialogues per person.

\subsection{Participant Information}
\label{sec:participant}
Prior to recording our dataset, we received an IRB approval to collect facial video, speech, and text data to build human multimodal dialogue technology. We recruited students at a university who were fluent in English and could fulfill the designated portion of the dialogues.  A recruitment notice included general information about TopicalChat, the dataset to be recorded, wage and responsibilities of the participants, and potential effects and contributions of building a multimodal dialogue dataset. After receiving 25 applications, interviews were conducted on all applicants. During the interview, we notified that we will be collecting audiovisual data of the participant during recording sessions, which will be released to the research field in the future.  We also collected participant information such as race, sex, nationality and age, agreement to release audiovisual data, and assessed the English fluency and ability to read and act out a given dialogue script with emotions. Two interviewees in charge of the dataset collection selected actors by ranking each participant on a scale of 1 to 5 on each criterion and considering the diversity of participant demographics. Thus, six female and six male actors from six different countries, and age varying from 20 to 30 were selected. Details on participant information are outlined in Table~\ref{tab:participant_info}.

\begin{table}
    \renewcommand{\arraystretch}{1.2}
    \renewcommand{\tabcolsep}{2.1mm}
    \centering
    \resizebox{0.99\linewidth}{!}{
    \begin{tabular}{lccccccc}
        \Xhline{3\arrayrulewidth}
        \textbf{Id} & \textbf{Gender} & \textbf{Age} & \textbf{Nationality} & \textbf{\# dialogues} & \textbf{Acc.}\\
        \midrule
        a & F & 24 & Indonesia & 1,453 & 69.3\\
        b & F & 25 & S. Korea & 1,454 & 63.6\\
        c & M & 23 & Kazakhstan & 1,772 & 59.3\\
        d & M & 23 & Kazakhstan & 1,108 & 33.8\\
        e & F & 24 & India & 1,718 & 41.5\\
        f & M & 24 & Pakistan & 1,083 & 43.8 \\
        g & F & 20 & Kazakhstan & 1,774 & 50.0 \\
        h & M & 21 & Pakistan & 1,642 & 37.0\\
        i & F & 23 & Pakistan & 995 & 60.0\\
        j & M & 24 & Bangladesh & 1,661 & 44.7\\
        k & M & 20 & S. Korea & 1,449 & 44.0\\
        l & F & 20 & Pakistan & 1,357 & 21.2\\
        \bottomrule
        \Xhline{3\arrayrulewidth}
    \end{tabular}}
    \vspace{-0.1cm}
    \caption{Participant information of MultiDialog.}
    \label{tab:participant_info}
    \vspace{-0.5cm}
\end{table}

After all participants were selected, we held an orientation to guide participants on the recording procedure. For a single recording session of three hours, two participants were scheduled to film 50 to 60 conversations in TopicalChat. The number of conversations to film in a session was calculated based on a trial recording session, in which two speakers filmed approximately 60 conversations in a three-hour period, including breaks. Participants learned how to navigate through the dialogue display program to start and end recording conversations, and proceed to the next utterance. The display program showed the conversation script along with the corresponding emotion for each utterance, and the remaining number of conversations to film in the current session. We notified each participant to attach a microphone about 15 to 20 cm from their mouth and adjust the camera to the shoulder level before recording. Lastly, we collected consent forms for providing personal information for compensation and informed consent forms for human subject research participants.










\subsection{Annotation Evaluation}
We conducted a comprehensive user study involving 25 participants, where we randomly sampled 70 utterances from the dataset and participants predicted the emotions conveyed within each utterance to verify the quality of emotions. 

Table~\ref{tab:participant_info} includes the accuracy of each actor in conveying the intended emotion in the utterance. Given that real-life conversations often involve subtle and layered emotional expressions, the dataset was designed to mirror this intricacy. Based on previous research \cite{hu2018deep, wang2021local} on subtle emotion recognition, the results from our user study underscore the effectiveness of the actors in portraying these subtle emotions. To enhance the quality of the emotion annotations to be used in future research, we filter out recordings from actors that exhibit low prediction scores and release a subset of MultiDialog. 

\begin{table}
    \renewcommand{\arraystretch}{1.3}
    \renewcommand{\tabcolsep}{2.1mm}
    \centering
    \resizebox{0.99\linewidth}{!}{
    \begin{tabular}{lccccccc}
        \Xhline{3\arrayrulewidth}
        \textbf{} & \textbf{NEU} & \textbf{HAP} & \textbf{FEAR} & \textbf{ANG} & \textbf{DISG} & \textbf{SUR} & \textbf{SAD} \\
        \hline
        \textbf{NEU}    & 0.88 & 0.04 & 0.01 & 0.03 & 0.00 & 0.03 & 0.02 \\
        \textbf{HAP}      & 0.18 & 0.75 & 0.00 & 0.01 & 0.01 & 0.05 & 0.001 \\
        \textbf{FEAR}    & 0.09 & 0.02 & 0.39 & 0.03 & 0.13 & 0.22 & 0.13 \\
        \textbf{ANG}      & 0.07 & 0.00 & 0.07 & 0.76 & 0.14 & 0.02 & 0.00 \\
        \textbf{DISG}  & 0.02 & 0.00 & 0.02 & 0.11 & 0.83 & 0.02 & 0.00 \\
        \textbf{SUR}  & 0.14 & 0.13 & 0.00 & 0.04 & 0.01 & 0.68 & 0.00 \\
        \textbf{SAD}        & 0.12 & 0.00 & 0.14 & 0.04 & 0.10 & 0.00 & 0.59 \\
        \bottomrule
        \Xhline{3\arrayrulewidth}
    \end{tabular}}
    \vspace{-0.1cm}
    \caption{Confusion matrix of emotion categories estimated from the user study.}
    \label{table:confusion_matrix}
    \vspace{-0.5cm}
\end{table}

Table~\ref{table:confusion_matrix} is the confusion matrix between emotion categories estimated from the user study, focusing on results from actors who achieved above 40\% emotion accuracy. The result closely aligns with the human innate ability to recognize emotion from audio-visual \cite{busso2008iemocap}, underlining the effectiveness of MultiDialog in conveying emotion within utterances. Certain emotions, such as fearful and sad, exhibited lower accuracy rates, which we attribute to the inherent complexity and subtlety of these emotions in natural conversations \cite{poria2018meld}. 

\subsubsection{Gold Emotion Dialogue Subset}
\label{sec:goldemotion}
We provide a gold emotion dialogue subset in the MultiDialog dataset, a more reliable resource for studying emotional dynamics in conversations. Previous research \cite{hu2018deep, wang2021local} indicates that the accuracy rates for recognizing subtle emotions are slightly under 40\%. Thus, we classify dialogues from actors that exhibit emotion accuracy above 40\% as gold emotion dialogue. We release the gold emotion annotations of actor IDs along with the dataset in \url{https://huggingface.co/datasets/IVLLab/MultiDialog}.

\subsection{Recording Setup}

\label{sec:dataset1}
\begin{figure}[h]
	\begin{minipage}[b]{\linewidth}
		\centering		\centerline{\includegraphics[width=7cm]{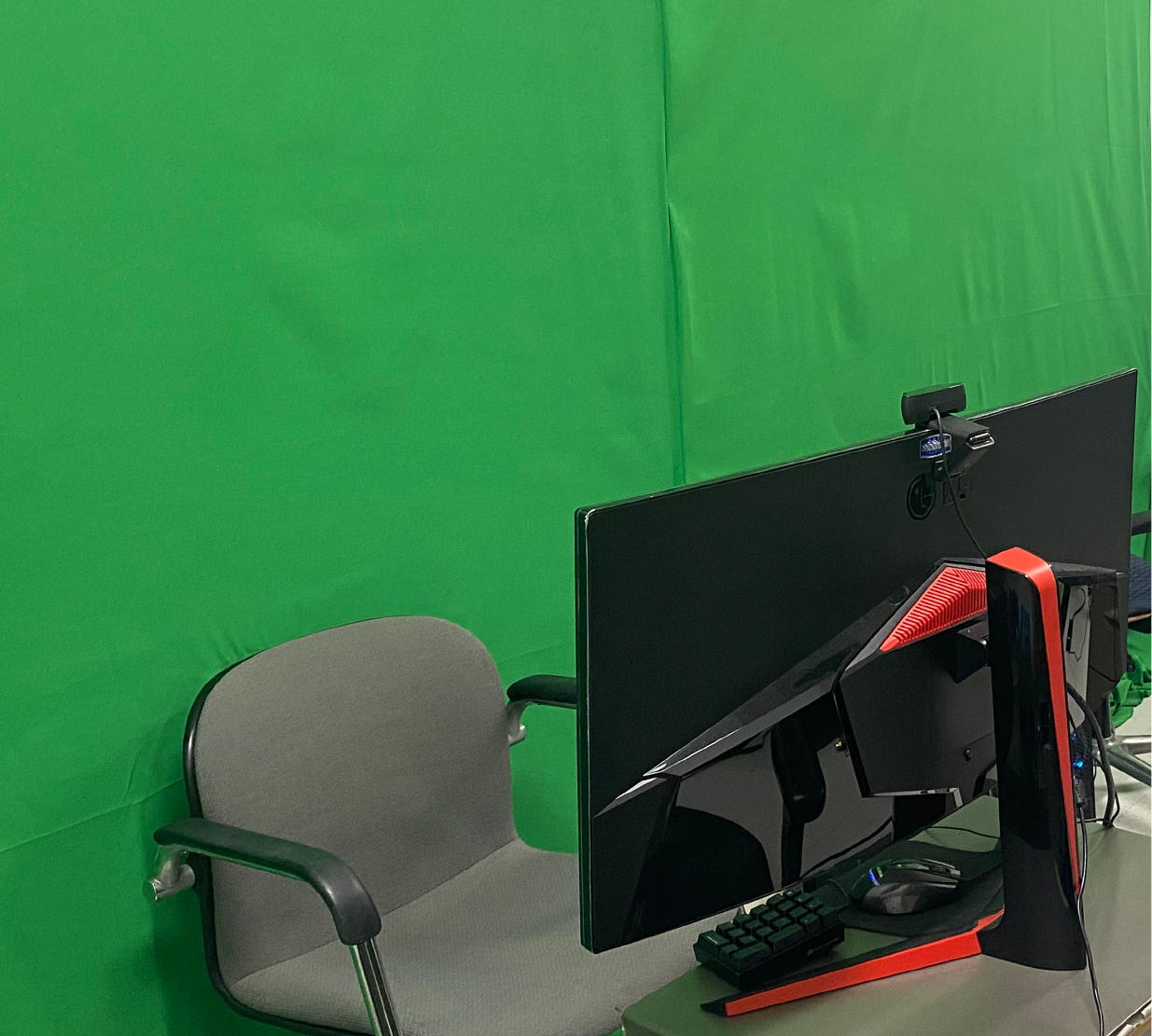}}
	\end{minipage}
	\caption{Recording studio setup for MultiDialog dataset}
	\label{fig:5}
  \vspace{-0.5cm}
\end{figure}

Fig.~\ref{fig:5} shows the studio setup for recording sessions.

\section{Evaluation Metrics}
\noindent \textbf{BLEU} \cite{post2018call} evaluates the fluency and adequacy of generated responses based on n-gram overlap. A higher BLEU score indicates a more natural and engaging dialogue model.

\noindent \textbf{PPL} \cite{bengio2000neural} measures how well a language model predicts the generated response. A lower perplexity indicates that the model is more confident and accurate in predicting the next word, suggesting higher quality in generating coherent and contextually relevant responses.

\noindent \textbf{DISTINCT-n} \cite{li-etal-2016-diversity} evaluates the diversity of generated response by calculating the percentage of unique n-grams in the set of responses. Specifically, D-1 measures the percentage of unique unigrams in the generated text, while D-2 measures the percentage of unique bigrams. 

\noindent \textbf{METEOR} \cite{banerjee2005meteor} (Metric for Evaluation of Translation with Explicit Ordering) evaluates the quality of generated response by computing the alignment-based precision and recall between the generated output and the ground truth, considering synonyms and paraphrases. 

\noindent \textbf{F1} \cite{banerjee2005meteor} combines the accuracy of the generated response (precision) and the coverage of the relevant response (recall). It provides a balanced measure of how well the model performs in generating relevant and accurate responses.

\end{document}